\pdfoutput=1
\PassOptionsToPackage{table}{xcolor}

\documentclass[11pt]{article}

\usepackage[]{acl}

\usepackage{times}
\usepackage{latexsym}
\usepackage{booktabs}
\usepackage{multirow}
\usepackage{graphicx}
\usepackage{amsmath}
\usepackage{amsfonts}
\usepackage{xcolor}
\usepackage{todonotes}
\usepackage{listings}
\usepackage{subcaption}
\usepackage[T1]{fontenc}
\usepackage[utf8]{inputenc}

\usepackage{microtype}

\usepackage{inconsolata}

\usepackage{CJKutf8}  % xrliu: Chinese in the appendix

\title{LongWanjuan: Towards Systematic Measurement for Long Text Quality}

\author{
Kai Lv\textsuperscript{1,2}\thanks{\ \ Equal contribution. }, 
Xiaoran Liu\textsuperscript{1,2}\footnotemark[1],  
Qipeng Guo\textsuperscript{2}, 
Hang Yan\textsuperscript{2}, 
Conghui He\textsuperscript{2},\\
\textbf{Xipeng Qiu\textsuperscript{1}, 
Dahua Lin\textsuperscript{2}}\\
\textsuperscript{1}School of Computer Science, Fudan University, \textsuperscript{2}Shanghai AI Laboratory\\
\texttt{\{klv21,liuxr22\}@m.fudan.edu.cn},\\ \texttt{\{guoqipeng,yanhang,heconghui,lindahua\}@pjlab.org.cn}\\
\texttt{xpqiu@fudan.edu.cn}\\
}

\begin{document}
\maketitle
\begin{abstract}
The quality of training data are crucial for enhancing the long-text capabilities of foundation models. Despite existing efforts to refine data quality through heuristic rules and evaluations based on data diversity and difficulty, there's a lack of systematic approaches specifically tailored for assessing long texts. 
Addressing this gap, our work systematically measures the quality of long texts by evaluating three fundamental linguistic dimensions: coherence, cohesion, and complexity.
Drawing inspiration from the aforementioned three dimensions, we introduce a suite of metrics designed to evaluate the quality of long texts, encompassing both statistical and pre-trained language model-based ones.
Leveraging these metrics, we present LongWanjuan, a bilingual dataset specifically tailored to enhance the training of language models for long-text tasks with over 160B tokens. In LongWanjuan, we categorize long texts into holistic, aggregated, and chaotic types, enabling a detailed analysis of long-text quality. 
Furthermore, we devise a data mixture recipe that strategically balances different types of long texts within LongWanjuan, leading to significant improvements in model performance on long-text tasks. The code and dataset are available at \url{https://github.com/OpenLMLab/LongWanjuan}.

\end{abstract}

\section{Introduction}

Effectively processing long texts is a crucial capability of language models and has recently become a focal point of research~\cite{ScalingRope,yarn,Giraffe,LM-Infinite,pi}. Tasks such as long document summarization~\cite{QMSum}, long document question answering~\cite{Qasper}, repository-level code tasks~\cite{RepoBench}, and retrieval-augmentation generation~\cite{rag} often involve handling thousands or even tens of thousands of tokens. 

\begin{figure}
    \centering
    \includegraphics[width=0.86\linewidth]{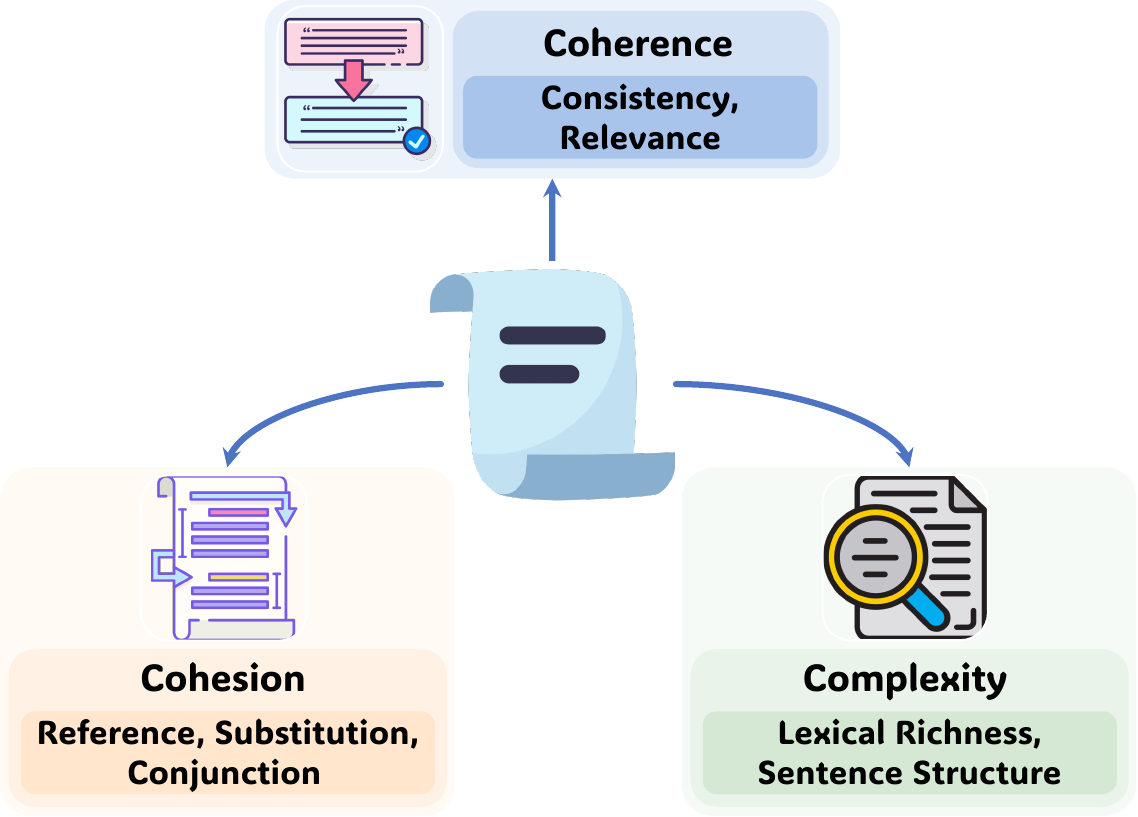}
    \caption{The three dimensions for measuring the quality of long texts: coherence, cohesion and complexity.}
    \label{fig:3d}
\end{figure}

The quality of data is vital for the long-text capabilities of foundation models~\cite{Data-centric-survey,llama2long,codellama}. 
There have been efforts made to improve data quality. Some approaches employ heuristic rules, such as deduplication and the removal of overly short data entries~\cite{slimpajama,refinedweb}. Additionally, some other approaches  consider data diversity and perplexity based on pre-trained language models~\cite{D4,WhenLessisMore}. However, these filtering rules are designed for general training data and do not take into account the unique characteristics of long texts.

To systematically assess the quality of long texts, we adhere to linguistic fundamentals and evaluate them through three dimensions: coherence~\cite{coherence}, cohesion~\cite{cohesion,Cohesionisnotcoherence}, and complexity~\cite{complexity}, as illustrated in Figure~\ref{fig:3d}. 
Given that long texts typically contain more extensive content, they necessitate elevated levels of these characteristics to effectively convey information and engage in discussion.
\textbf{Coherence} measures the overall consistency and clarity of the text as a whole. \textbf{Cohesion} gauges the strength of connections between sentences or sections of the text. \textbf{Complexity} assesses the linguistic sophistication within the text. 
Drawing from these three fundamental dimensions, we propose a set of metrics to quantitatively analyze the quality of long texts. These metrics encompass both statistical and pre-trained model-based approaches, offering strong interpretability. Further details on these metrics can be found in Section~\ref{sec:method}.

Based on the characteristics across these three dimensions, we categorize the long texts in pre-training dataset into three types: \textbf{holistic long texts}, encompassing complete works such as books, academic papers, reports, novels, and interviews; \textbf{aggregated long texts}, consisting of short texts related by topic or fragmented texts like extensive lists or tables; and \textbf{chaotic long texts}, characterized by nonsensical content such as garbled data. 
Drawing upon these classifications, we manually annotated a validation set of 200 samples from SlimPajama~\cite{slimpajama} and Wanjuan~\cite{he2023wanjuan} to validate the correlation between our proposed metrics and human judgments. Our quantitative metrics effectively differentiate between the three categories of long texts.

Building on these analysis and metrics, we create a bilingual long-text dataset with category labels, named LongWanjuan, containing over 160B tokens.
With LongWanjuan, we propose a data mixture recipe to mitigate the imbalance between holistic long texts and aggregated long texts within the dataset. 
Specifically, by removing chaotic long texts and upsampling aggregated long texts, we continue to train InternLM2-7B~\cite{internlm} with an additional 5B tokens, thereby achieving state-of-the-art performance for long texts on models of the 7B parameter scale.
The effectiveness and generalizability of this recipe are analyzed in Section~\ref{sec:ablation}.

In summary, our contributions are as follows: 
\begin{enumerate}
    \item To the best of our knowledge, this is the first work to systematically analyze and introduce quantitative metrics for assessing the quality of long texts. Grounded in linguistic principles, we measure the quality of long texts in terms of coherence, cohesion, and complexity. 
    \item Leveraging SlimPajama and Wanjuan, we constructed a bilingual long-text dataset with over 160B tokens, LongWanjuan, which is available to the community as an open-source resource. 
    \item Based on LongWanjuan, we devise a data mixture recipe to mitigate the imbalance in the dataset, and advance to a new state-of-the-art long-text model at the 7B parameter scale, demonstrating a 13.07\% improvement over the untrained baseline on Longbench~\cite{bai2023longbench}.
\end{enumerate}

\section{Related Work}
\subsection{Pre-training Data Pruning}
The quality of pre-training data plays a crucial role in the performance of foundation models~\cite{gopher,glam,llama2long,codellama,phi}. Several studies have enhanced data quality by pruning the original training data into a subset. 

Some works primarily focus on heuristic rules and deduplication to improve data quality. 
\citet{t5} and \citet{slimpajama} employ similar heuristic rules to enhance data quality, including the removal of overly short entries and deduplication.
~\citet{Semdedup} leverages embeddings from pre-trained models to further eliminate semantic duplicates.
Another notable contribution is RefinedWeb~\cite{refinedweb}, which meticulously designs a comprehensive data processing pipeline.
% , including extracting the core content of web pages, applying multi-scale filtering operations.

Moreover, several studies take into consideration the data diversity and difficulty to prune data. 
\citet{D4} employs clustering-based methods to augment data diversity.
% , which initially deduplicates data within clusters, and subsequently removes samples proximal to the centroid. 
\citet{WhenLessisMore} evaluates the effectiveness of perplexity, EL2N~\cite{el2n}, and memorization score~\cite{MemorizationScore} in assessing data difficulty.
% , finding that even the most basic metric, perplexity, can outperform more computationally intensive scoring methods. 
~\citet{D2} regards data diversity and difficulty as complementary aspects, selecting data through forward and reverse message passing on a dataset graph.

Distinct from these studies that concentrate on general pre-training data, our research specifically targets long texts. 
It is essential to highlight that our work extends beyond mere data curation and is applicable in a wider range of contexts for evaluating the quality of long texts.

\subsection{Text Quality Assessment}
\begin{table*}[t]
\small
\centering
\begin{tabular}{@{}p{1.6cm}p{5.9cm}p{7.4cm}@{}}
\toprule
 & \textbf{Low Level Example}      & \textbf{High Level Example}       \\ \midrule
\textbf{Coherence} & The project aims to reduce carbon emissions by 25\% within the next five years. Strawberries are rich in vitamins and antioxidants. It's raining today. & The project \textbf{\textcolor{blue}{aims to}} \textbf{\textcolor{orange}{reduce carbon emissions}} by 25\% within the next five years. \textbf{\textcolor{blue}{This goal}} will be achieved through the implementation of \textbf{\textcolor{orange}{renewable energy sources}} and improved \textbf{\textcolor{orange}{energy efficiency}}. \textbf{\textcolor{blue}{The initiative}} reflects our commitment to \textbf{\textcolor{orange}{environmental sustainability}}. \\ \hline
\textbf{Cohesion} & I prepared the soil in my garden. I planted some tomato seeds. I watered seeds in my garden. & \textbf{\textcolor{blue}{Firstly}}, I prepared the soil in my garden. \textbf{\textcolor{blue}{Then}}, I planted some tomato seeds in \textbf{\textcolor{orange}{the prepared ground}}. \textbf{\textcolor{blue}{After that}}, I watered \textbf{\textcolor{orange}{them}}. \\ \hline
\textbf{Complexity} & Eating fish is good. It helps your brain. &  After researching various \textbf{\textcolor{blue}{nutrition sources}}, I concluded \textbf{\textcolor{orange}{that}} incorporating \textbf{\textcolor{blue}{omega-3 fatty acids}} and \textbf{\textcolor{blue}{antioxidants}} into our diet can significantly \textbf{\textcolor{blue}{ameliorate cognitive decline}} in elderly individuals.  \\ \bottomrule
% Metric & PPL, ACC & conn proun dmv & ttr len of paragraph \\ 
\end{tabular}
\caption{Examples illustrating dimensions of coherence, cohesion, and complexity. \textcolor{blue}{Blue} and \textcolor{orange}{orange} illustrate distinct aspects of each dimension. In the context of coherence, the \textcolor{blue}{blue} and \textcolor{orange}{orange} texts signify different elements that maintain thematic consistency throughout the text. For cohesion, the \textcolor{blue}{blue} text indicates connectors that link sentences together, while the \textcolor{orange}{orange} text refers to references to previously mentioned entities. Within complexity, the \textcolor{blue}{blue} text represents lexical sophistication, whereas the \textcolor{orange}{orange} text denotes the complexity of sentence structure.}
\label{tab:examlpes}
\end{table*}

Several works score texts through supervised learning. 
\citet{TextScoring} trains score-specific word embeddings and a Long Short-Term Memory (LSTM) network~\cite{lstm} for text scoring purposes. Similarly, \citet{cn_essay} conducts fine-grained annotations on 501 Chinese essays and achieves comparable scoring performance to ChatGPT-3.5 through training based on RoBERTa~\cite{roberta}. However, these approaches suffer from limited generalizability, being applicable only within the confines of labeled domains.

Other works leverage unsupervised methods to automatically construct data for training purposes. 
UNION~\cite{union}
% , built upon BERT~\cite{bert}, 
is trained to differentiate between human-written stories and negative samples. 
\citet{dmr} explores implicit discourse relations with a latent discourse sense, showcasing strong performance.

Furthermore, some studies utilize pre-trained language models to assess text quality without additional training. 
\citet{Coherence_wo_sup} evaluates textual coherence by modeling the uncertainty of topics within paragraphs and their interrelations, thus scoring texts. 
BARTScore~\cite{BARTScore} and GPTScore~\cite{GPTScore} employ the weighted average of the model’s output conditional probabilities as a metric, facilitating multifaceted evaluation across a broad range of generative tasks.

Our work measures the quality of long texts from multiple dimensions, introducing metrics that are task-agnostic and do not necessitate additional training.

\section{Method}
\label{sec:method}
% A Three-Facet Measurement for Long Text Quality
Long texts, characterized by their extended contexts and abundant information, pose distinct challenges in maintaining textual integrity and quality.
We systematically measure the quality of long texts through three dimensions: coherence, cohesion, and complexity. 
Each dimension is accompanied by corresponding quantitative metrics, allowing for an effective measurement of long text quality.

\subsection{Coherence, Cohesion and Complexity}

In accordance with linguistic fundamentals, we systematically assess the quality of long texts through the following three dimensions.

\textbf{Coherence} refers to the consistency and clarity of the text as a whole. 
A coherent text maintains thematic unity throughout its parts, with logical connections between the different sections. 

\textbf{Cohesion} measures the degree of tight connection between two sentences or sections of the text, reflected in the use of connectives, pronouns, synonyms, and hypernyms/hyponyms.

\textbf{Complexity} assesses the level of linguistic sophistication in the use of language in the text. This can be gauged through the richness and diversity of vocabulary, as well as the complexity of sentence structures. 

To better elucidate these dimensions, we provide examples in Table~\ref{tab:examlpes} that illustrate both high and low levels of these dimensions. Key terms that exemplify specific features of each dimension are highlighted for emphasis.

\subsection{Metric}
Inspired by the three dimensions mentioned above, we propose the following metrics to assess the quality of long text $\boldsymbol{t}=\{t_1, t_2, \ldots, t_n\}$, including both statistical and model-based ones, where higher values correlate with more pronounced characteristics of the corresponding dimension.

\begin{figure*}[t]
    \centering
    \includegraphics[width=1\linewidth]{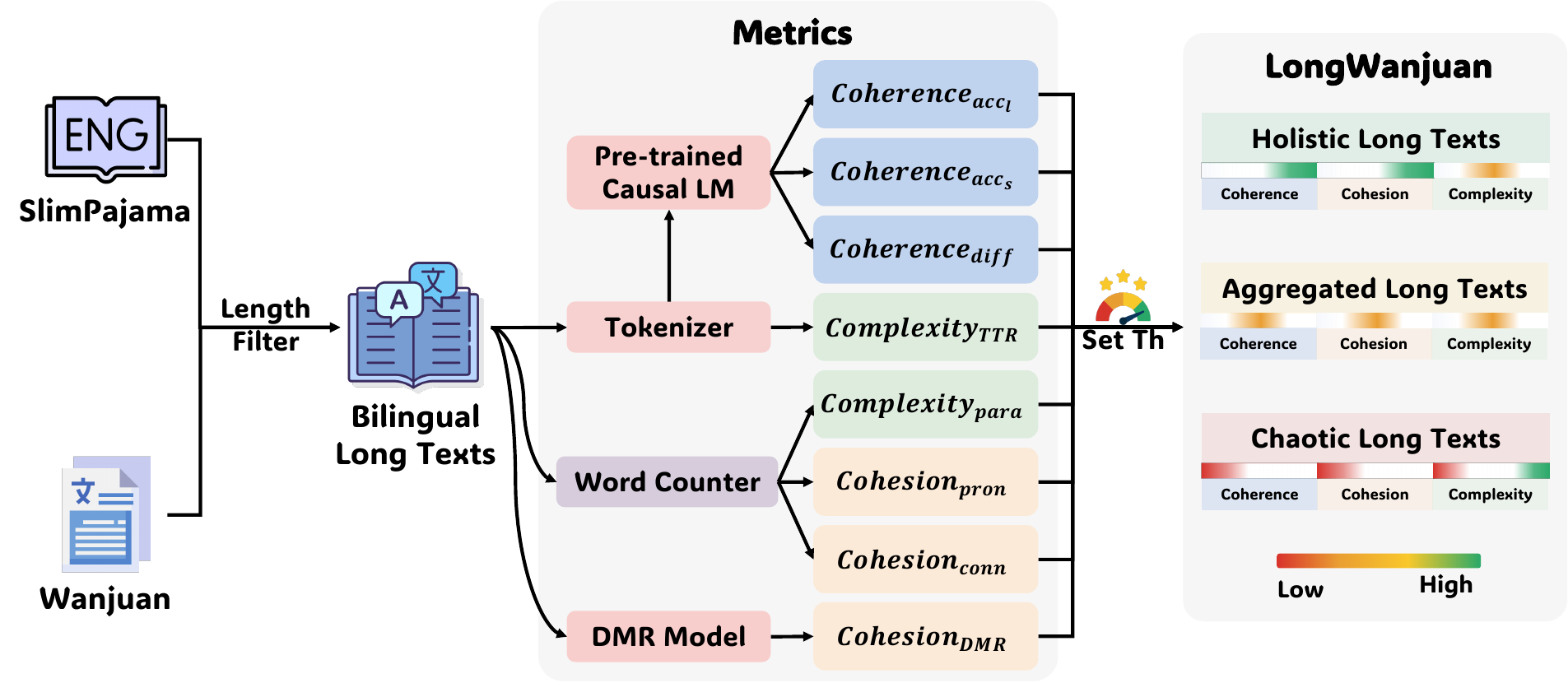}
    \caption{Pipeline for constructing the LongWanjuan dataset.}
    \label{fig:pipeline}
\end{figure*}

To measure the coherence of a long text, we evaluate the extent to which prior segments of the text contribute to understanding subsequent segments. 
A coherent text should make it easier to predict its following content based on its preceding context. 
For example, when predicting the blue text below, it is easier to make a correct prediction if the preceding text is provided.
% \begin{figure}[t]
    % \centering
\begin{center}
    \includegraphics[width=0.85\linewidth]{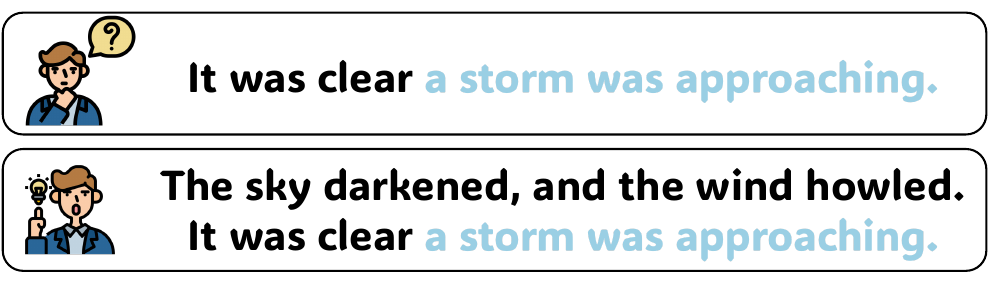}
    % \caption{Enter Caption}
    \label{fig:enter-label}
\end{center}
    
We evaluate the coherence of long texts by comparing the prediction accuracy with a longer context and the accuracy with a shorter context, as well as the difference between these two contexts. 
Specifically, with a pre-trained causal language model parameterized by $\theta$, we employ the following three metrics for assessing the coherence of long texts:

\begin{align}
\text{Coherence}_{\text{acc}_l} &= \sum_{i=1}^{\left\lfloor\frac{n}{w}\right\rfloor} acc\left(\boldsymbol{y}^i|\boldsymbol{x}_l^i,\theta\right) / \left\lfloor\frac{n}{w}\right\rfloor ,\\
\text{Coherence}_{\text{acc}_s} &= \sum_{i=1}^{\left\lfloor\frac{n}{w}\right\rfloor} acc\left(\boldsymbol{y}^i|\boldsymbol{x}_s^i,\theta\right) / \left\lfloor\frac{n}{w}\right\rfloor , \\
\text{Coherence}_{\text{diff}} &= \frac{\sum_{i=1}^{\left\lfloor\frac{n}{w}\right\rfloor} \frac{\ell\left(\boldsymbol{y}^i|\boldsymbol{x}_l^i,\theta\right)-\ell\left(\boldsymbol{y}^i|\boldsymbol{x}_s^i,\theta\right)}{\ell\left(\boldsymbol{y}^i|\boldsymbol{x}_l^i,\theta\right)}}{ \left\lfloor\frac{n}{w}\right\rfloor} , \\
% \boldsymbol{y}^i &= \boldsymbol{t}_{(i-\frac{1}{4})w:iw} &, \notag\\
% \boldsymbol{x}_l^i &= \boldsymbol{t}_{(i-1)w:(i-\frac{1}{4})w} &, \notag\\
% \boldsymbol{x}_s^i & = \boldsymbol{t}_{(i-\frac{1}{2})w:(i-\frac{1}{4})w} &. \notag \\
% \text{where } \boldsymbol{y}^i = \{t&_{(i-\frac{1}{4})w}, t_{(i-\frac{1}{4})w+1}, \ldots, t_{iw}\} &, \notag\\
% \boldsymbol{x}_l^i = \{t&_{(i-1)w}, t_{(i-1)w+1}, \ldots, t_{(i-\frac{1}{4})w}\} &, \notag\\
% \boldsymbol{x}_s^i  = \{t&_{(i-\frac{1}{2})w}, t_{(i-\frac{1}{2})w+1}, \ldots, t_{(i-\frac{1}{4})w}\} &. \notag
\text{where } \boldsymbol{x}_l^i &= \{t_{(i-1)w},  \ldots, t_{(i-\frac{1}{4})w}\} , \notag\\
\boldsymbol{x}_s^i  &= \{t_{(i-\frac{1}{2})w},  \ldots, t_{(i-\frac{1}{4})w}\} , \notag \\
\boldsymbol{y}^i &= \{t_{(i-\frac{1}{4})w},\ldots, t_{iw}\} .
\end{align}
$acc(\boldsymbol{y}|\boldsymbol{x},\theta)$ and $\ell(\boldsymbol{y}|\boldsymbol{x},\theta)$ denote the model's average top-1 prediction accuracy and negative log-likelihood loss for generating $\boldsymbol{y}$ given the prompt $\boldsymbol{x}$, parameterized by $\theta$.
$\text{Coherence}_{\text{acc}_l}$ and $\text{Coherence}_{\text{acc}_s}$ respectively denote the model's top-1 prediction accuracy with longer and shorter preceding texts, and $\text{Coherence}_\text{diff}$ represents the proportional improvement in model performance when using a longer versus a shorter context.
We process long texts with a sliding window of size $w$ to avoid exceeding the processing capabilities of the language model, setting $w$ to 4096 in practice. 

We quantitatively measure cohesion by analyzing the density of connectives and pronouns in the text and the relationships between adjacent sentences. 
Connectives play pivotal roles in linking words, sentences, or ideas within sentences and paragraphs. Pronouns, serving as substitutes for nouns or noun phrases, maintain references to specific entities mentioned earlier while avoiding unnecessary repetition. 
\begin{align}
\text{Cohesion}_{\text{conn}} &= \frac{N_\text{conn}}{n} ,\\
\text{Cohesion}_{\text{pron}} &= \frac{N_\text{pron}}{n} ,\\
\text{Cohesion}_{\text{DMR}} &= 1-\sum_{i=1}^{N}\frac{p(\text{no\_conn}|s_{i},s_{i+1})}{N} ,
\end{align}

where \(N_{\text{conn}}\) and \(N_{\text{pron}}\) represent the number of connectives and pronouns in the text, respectively. The comprehensive list of considered connectives and pronouns can be found in the Appendix~\ref{sec:conn}. The text \(\boldsymbol{t}\) consists of \(N+1\) sentences, with \(s_i\) denoting the \(i^{th}\) sentence in the text. The term \(p(\text{no\_conn}|s_i,s_{i+1})\) indicates the probability, as determined using Distributed Marker Representation (DMR)~\cite{dmr}, that sentences \(s_i\) and \(s_{i+1}\) are unrelated.\footnote{The DMR approach is originally considered for English texts only. To process Chinese data, we follow its training methodology and train a Chinese DMR model based on the Wanjuan dataset.}

The complexity of the text is assessed from vocabulary and paragraph.
\begin{align}
\text{Complexity}_{\text{TTR}} &= \frac{N_\text{unique}}{n} , \\
\text{Complexity}_{\text{para}} &= \frac{n}{N_\text{para}} ,
\end{align}
where \(N_{\text{unique}}\) refers to the number of unique tokens in the text, used to calculate the Type-Token Ratio (TTR)~\cite{ttr}. \(N_{\text{para}}\) denotes the number of paragraphs in the text, used to determine the average paragraph length.

\section{LongWanjuan}

\subsection{Dataset Construction}

Based on the analysis and metrics discussed previously, we introduce LongWanjuan, a bilingual long-text dataset. The pipeline for constructing our dataset is illustrated in Figure~\ref{fig:pipeline}.

Given that the majority of the SlimPajama~\cite{slimpajama} corpus is in English, we enrich it with Chinese texts from the Wanjuan~\cite{he2023wanjuan} dataset. 
Initially, we extract data entries exceeding 32K bytes from both the SlimPajama and Wanjuan datasets, serving as the starting point for our dataset construction. 

Subsequently, we evaluate each data entry using the metrics we proposed.
Specifically, we first tokenize the data with InternLM2 tokenizer~\cite{internlm}, thereafter calculating \(\text{Complexity}_{\text{TTR}}\). The tokenized results are further processed with InternLM2-7B to obtain coherence scores, including \(\text{Coherence}_{\text{acc}_l}\), \(\text{Coherence}_{\text{acc}_s}\), and \(\text{Coherence}_{\text{diff}}\). 
We employ NLTK~\cite{nltk} and LTP~\cite{LTP} respectively for English and Chinese sentence segmentation.
These sentences are then fed into DMR model to derive the \(\text{Cohesion}_{\text{DMR}}\) score. The metrics \(\text{Cohesion}_{\text{conn}}\), \(\text{Cohesion}_{\text{pron}}\) and \(\text{Complexity}_{\text{para}}\), are calculated by straightforward word counting.

\begin{figure}
    \centering
    \includegraphics[width=1\linewidth]{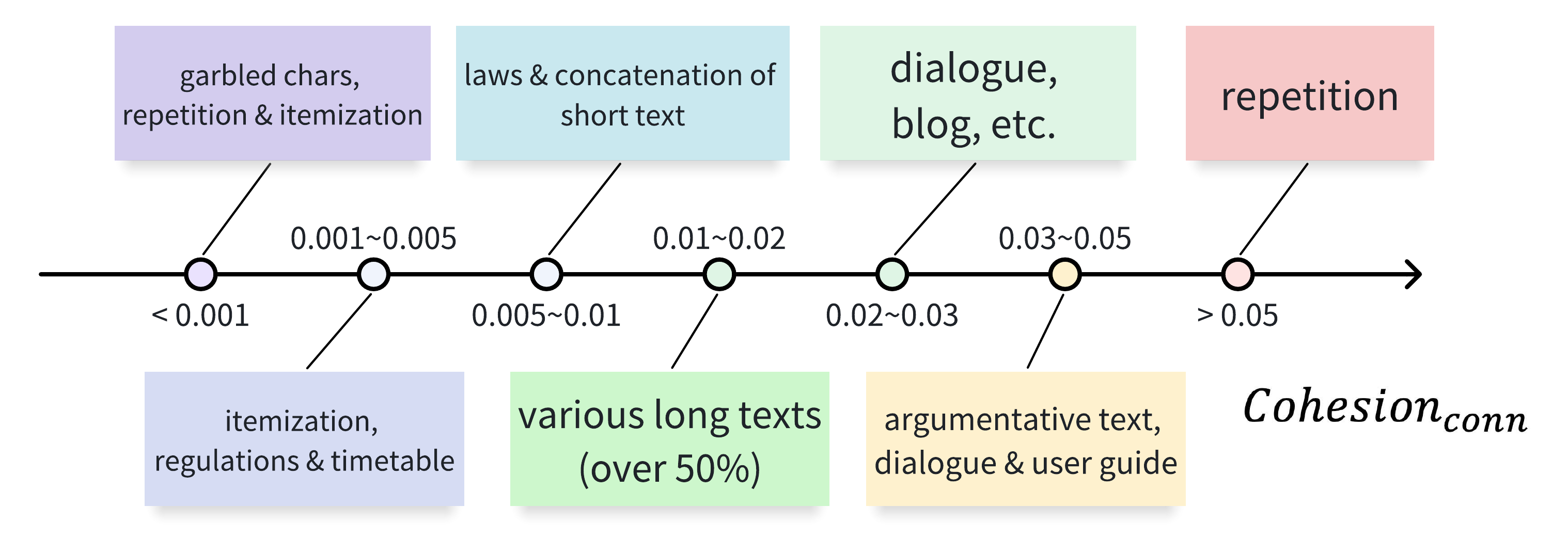}
    \caption{Distribution of texts with different characteristics on the \(\text{Cohesion}_\text{conn}\) metric in the C4 domain.}
    \label{fig:en_cohesion_conj}
\end{figure}

After scoring each data entry with these metrics, we establish thresholds to categorize the data into holistic long texts, aggregated long texts, and chaotic long texts. 
During this process, it is necessary only to check whether texts on either side of the threshold belong to different categories. Figure~\ref{fig:en_cohesion_conj} shows the distribution of texts within the C4 domain based on the \(\text{Cohesion}_{\text{conn}}\) metric. As illustrated, the texts within different ranges of our proposed metric exhibit distinct characteristics, simplifying the process of threshold determination. For each domain in the dataset, we can extract approximately 30 data samples based on the distribution of this metric and identify the thresholds between different categories of texts. More information on the distribution of text quality across various metrics are shown in Appendix~\ref{sec:detailed_stat}.
In this phase, we initially determine thresholds to segregate holistic long texts. Subsequently, within the remaining texts, we establish thresholds to differentiate chaotic long texts, with the residual texts classified as aggregated long texts.
% The thresholds employed are detailed in the Appendix D. 

Overall, holistic long texts are characterized by high coherence and cohesion, with moderate complexity. Aggregated long texts exhibit lower coherence and cohesion compared to the former. The main feature of chaotic long texts is their complexity, which is anomalously high or low.

\subsection{Statistics}
The LongWanjuan dataset comprises a total of 160.6B tokens, as tokenized by the InternLM2 tokenizer. Of these, holistic texts constitute 137.6B tokens, accounting for 85.7\% of the dataset; aggregated texts make up 21.8 billion tokens, or 13.6\%; and chaotic texts comprise 1.2B tokens, representing 0.7\%. In this section, we will present statistical information about LongWanjuan, focusing on the distribution of domains and lengths.
The specific values of token count and document count for each domain are provided in Appendix~\ref{sec:detailed_stat}.
\paragraph{Domain} 
\begin{figure*}[htbp]
    \centering
    \begin{subfigure}[b]{0.46\linewidth}
        \includegraphics[width=\linewidth]{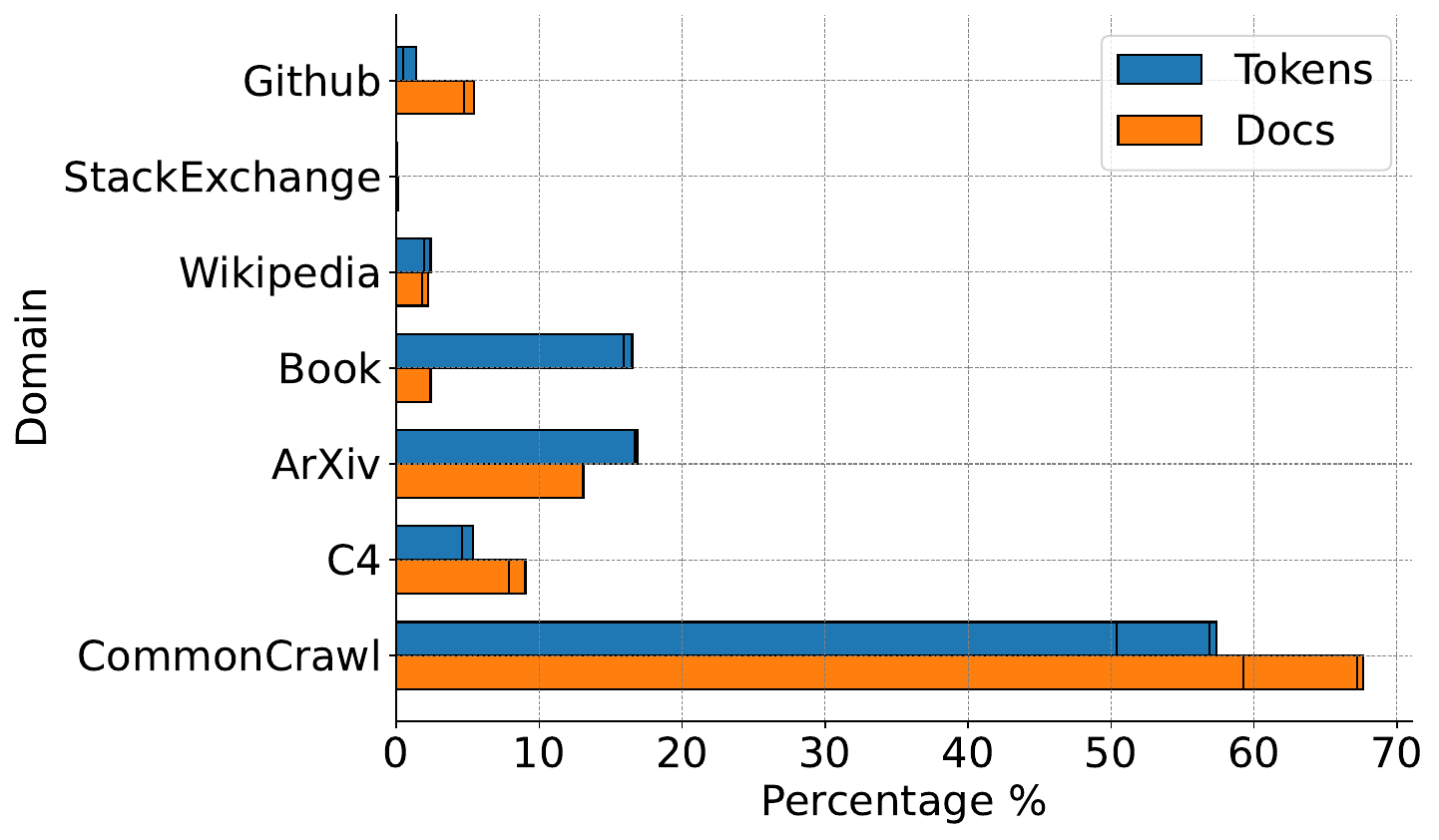}
        \caption{Distribution of data from SlimPajama.}
        \label{fig:stat_domain_en}
    \end{subfigure}
    \hfill
    \begin{subfigure}[b]{0.46\linewidth}
        \includegraphics[width=\linewidth]{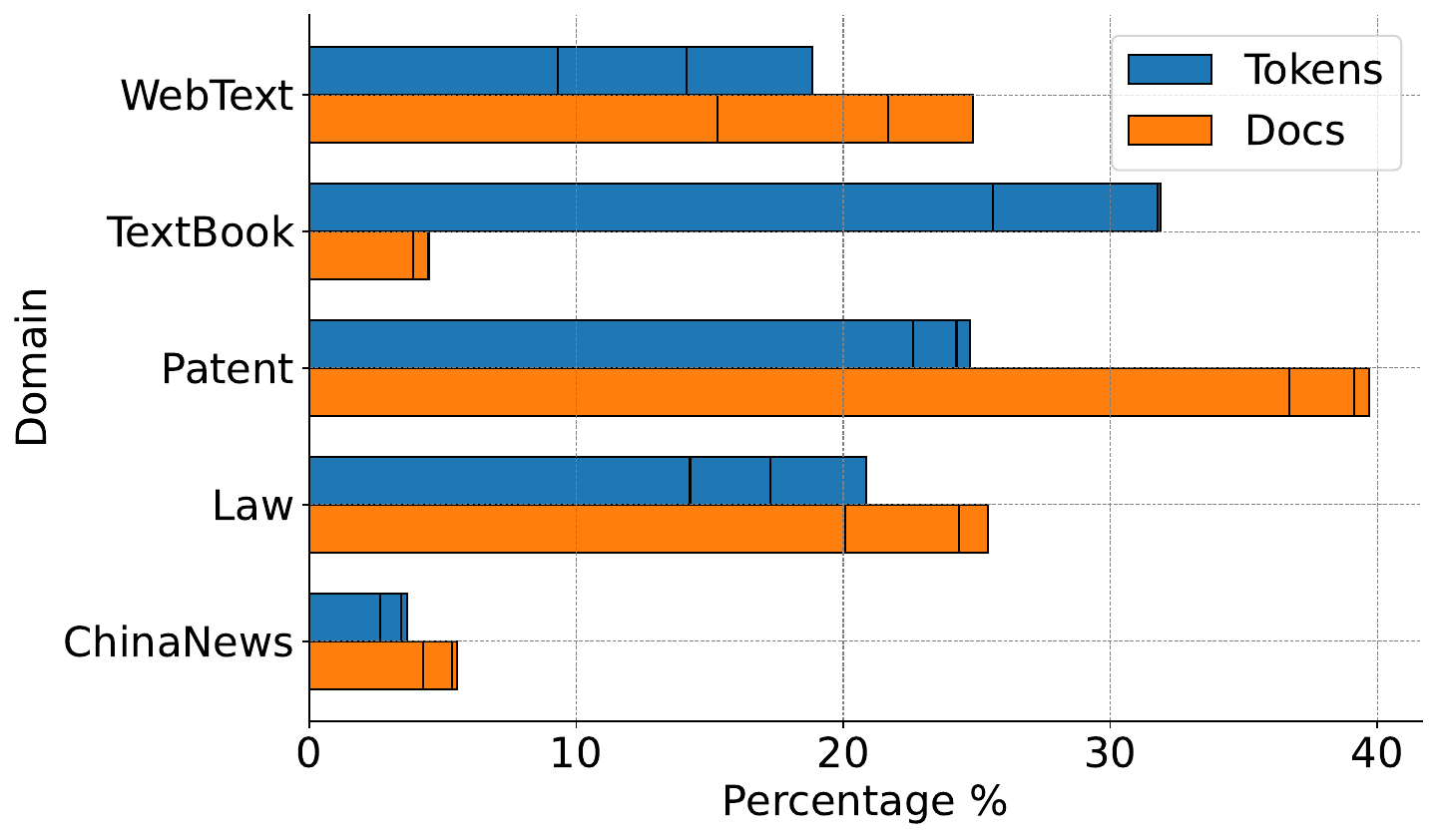}
        \caption{Distribution of data from Wanjuan.}
        \label{fig:stat_domain_cn}
    \end{subfigure}
    \caption{Distribution of token and document counts across different domains. Each bar is divided from left to right into three parts: holistic, aggregated, and chaotic texts.}
    \label{fig:mainlabel}
\end{figure*}

Figures~\ref{fig:stat_domain_en} and \ref{fig:stat_domain_cn} depict the distribution of data across various domains in English and Chinese, respectively, within the LongWanjuan dataset. In these bar graphs, each row is divided into three segments from left to right, representing holistic texts, aggregated texts, and chaotic texts, in that order. In the English data, the CommonCrawl domain predominates, accounting for over 50\% of the data. Apart from a significant amount of aggregated texts in the CommonCrawl domain, the majority of data in other domains consists of holistic texts. In the Chinese data, the distribution across different domains is more balanced, with each domain featuring both holistic and aggregated texts. The WebText and Law domains contain a notable number of chaotic texts. Detailed statistical information is available in Appendix~\ref{sec:detailed_stat}.

\paragraph{Length} 
\begin{figure}
    \centering
    \includegraphics[width=1\linewidth]{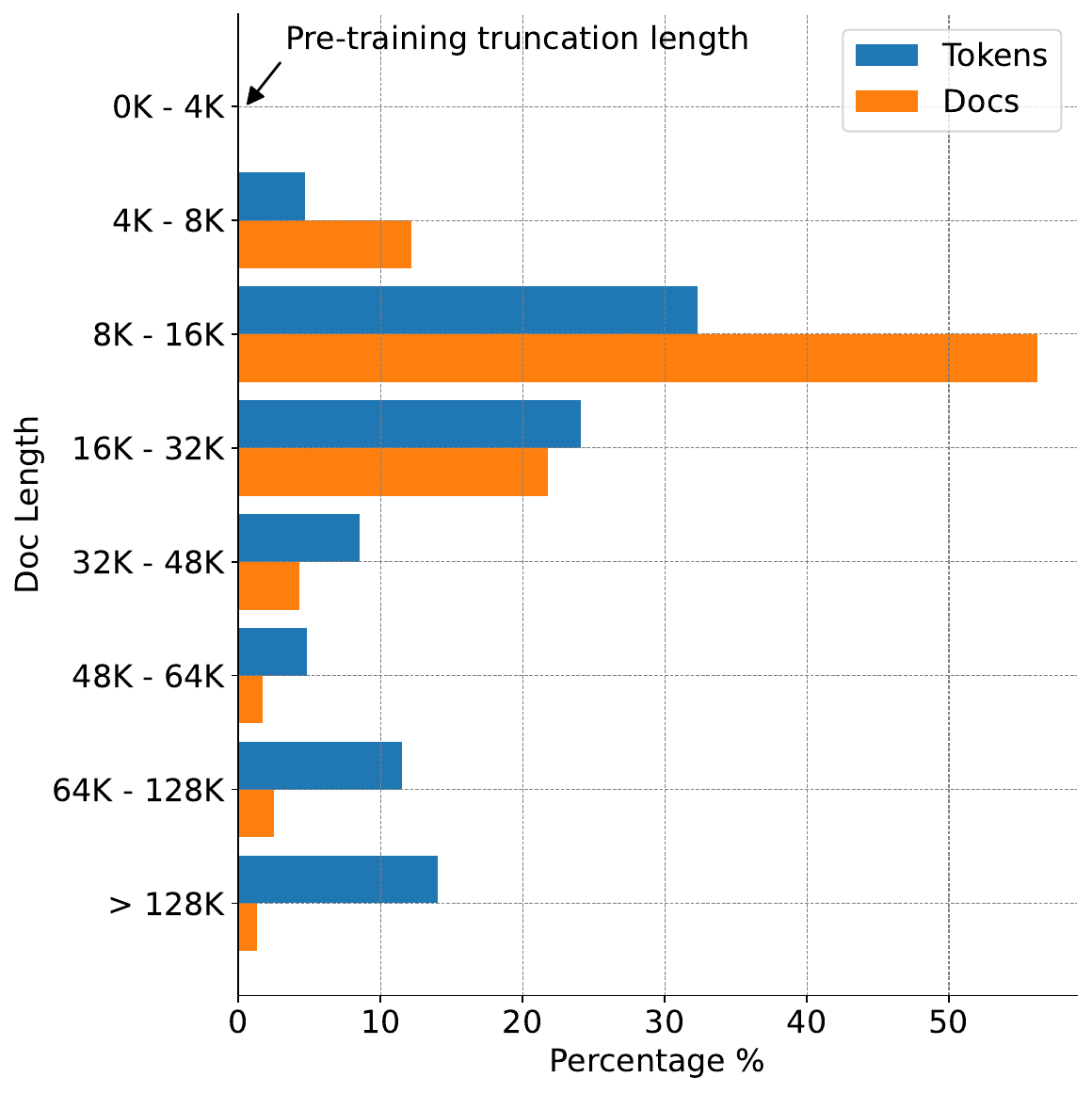}
    \caption{Distribution of token and document counts across different lengths. In LongWanjuan, over 99.9\% of the data exceed the truncation length in pre-training.}
    \label{fig:stat_len}
\end{figure}

Figure~\ref{fig:stat_len} illustrates the distribution of the number of data entries and the number of tokens across different lengths within the LongWanjuan dataset. During pre-training, the training data is generally truncated to a maximum length of 4K tokens, and entries of this length account for less than 0.1\% of the dataset in LongWanjuan. In terms of the number of tokens, more than 50\% of the data spans lengths between 8K and 32K tokens. Furthermore, over 10\% of the data exceeds a length of 128K tokens. With regard to the number of data entries, more than 50\% of the documents fall within the 8K to 16K token range. The trend in data entries by length initially increases before decreasing, and due to longer documents containing more tokens, the smallest quantity of tokens is observed in the 48K to 64K range.

\section{Experiments}

\begin{table}[!t]
\centering\small
\begin{tabular}{lcccc}
\toprule
 & \textbf{Holistic} & \textbf{Aggregated} & \textbf{Chaotic}  & \textbf{Total}\\ 
\midrule
EN & 0.97 & 0.87 & 0.81 & 0.91 \\
ZH & 0.97 & 0.58 & 0.79 & 0.80 \\
\bottomrule
\end{tabular}
\caption{The correlation between manual validation and the classification method we proposed}\label{tab_valid}
\end{table}
\begin{table*}[!t]
\centering\small
\begin{tabular}{llllll}
\toprule
& \textbf{EN}  & \textbf{ZH} & \textbf{Text} & \textbf{Code} & \textbf{Total} \\
\midrule
LongChat-v1.5-7B-32K &37.13 & 14.88 & 27.63 & 54.15 & 33.22\\
Yi-6B-200K & 37.65 & 15.12 & 28.04 & 64.55 & 35.72  \\
InternLM2-7B   & 51.61 & 34.07 & 40.91 & 62.86 & 45.43 \\ 
ChatGLM3-6B-32K    & 55.36 & \textbf{42.43} & 46.26 & 57.10 & 48.05 \\ 
\midrule
LLaMA2-7B with LongWanjuan  & 33.92 & 18.94 & 25.15 & 62.90 & 33.10 \\
InternLM2-7B with LongWanjuan & \textbf{56.64} & 39.31 & \textbf{46.26} & \textbf{65.26} & \textbf{50.26}\\
\bottomrule
\end{tabular}
\caption{Comparison between our proposed training strategy with other open-sourced LLMs on LongBench. The terms HOL, AGG, and CHA respectively denote holistic texts, aggregated texts, and chaotic texts.}\label{tab_main}
\end{table*}
\begin{table*}[!t]
\centering\small
\begin{tabular}{lcccccc}
\toprule
& \textbf{EN} & $\Delta$ & \textbf{ZH} & $\Delta$ & \textbf{Text} & $\Delta$ \\
\midrule
LLaMA2-7B-4K & 28.55 & & 13.62 & & 21.41 & \\ 
HOL. + AGG. + CHA.    & 32.86 &{\footnotesize +15.11\%} & 17.18 &{\footnotesize +26.20\%} & 24.30 &{\footnotesize +13.46\%} \\ 
HOL.   & 33.17 &{\footnotesize +16.20\%} & 18.44 &{\footnotesize +35.44\%} & 24.63 &{\footnotesize +15.02\%} \\ 
HOL. + AGG. & 33.66 &{\footnotesize +17.91\%} & 17.14 &{\footnotesize +25.88\%} & 24.99 &{\footnotesize +16.70\%} \\  	 
HOL. + Upsampling AGG. & \textbf{33.92} &{\footnotesize +18.80\%} & \textbf{18.94} &{\footnotesize +39.09\%} & \textbf{25.15} &{\footnotesize +17.45\%} \\ 
\midrule
InternLM2-7B & 51.61 & & 34.07 & & 40.91 & \\ 
HOL. + AGG. + CHA.    & 55.03 &{\footnotesize +6.63\%} & 36.63 &{\footnotesize +7.52\%} & 44.49 &{\footnotesize +8.74\%} \\
HOL.   & 55.12 &{\footnotesize +6.81\%} & 36.97 &{\footnotesize +8.51\%} & 44.61 &{\footnotesize +9.04\%} \\
HOL. + AGG. & 55.54 &{\footnotesize +7.62\%} & 37.36 &{\footnotesize +9.67\%} & 44.79 &{\footnotesize +9.46\%} \\
HOL. + Upsampling AGG. & \textbf{56.64} &{\footnotesize +9.76\%} & \textbf{39.31} &{\footnotesize +15.38\%} & \textbf{46.26} &{\footnotesize +13.07\%} \\
\bottomrule
\end{tabular}
\caption{Comparison of different training strategies data on LongBench. We also report relative improvements over the pre-trained LLMs in the same way as LLaMA2Long~\citep{llama2long}. The terms HOL, AGG, and CHA respectively denote holistic texts, aggregated texts, and chaotic texts.}\label{tab_long}
\end{table*}

\subsection{Manual Validation}

Complementary to the following training and evaluating results, we conduct human validation by manually annotating the type of 200 long texts from SlimPajama~\citep{slimpajama} and Wanjuan~\citep{he2023wanjuan} and then calculating the classification accuracy. The verification set includes 120 items in English and 80 items in Chinese, covering various domains as well as all three types of long texts in SlimPajama and Wanjuan. The verification results are shown in Table \ref{tab_valid}.

The quantitative metrics we proposed can effectively distinguish the three types of long texts in SlimPajama and Wanjuan. Specifically, for Chinese, the accuracy of the aggregated long text is relatively low. This is because the `TextBook' domain in Wanjuan contains a large amount of classical Chinese texts, which have inherent differences compared to modern Chinese texts. On one hand, it is challenging for models and rule-based scoring methods to accurately distinguish between them. On the other hand, there exist difficulties and biases in human annotation of these data. As a result, the relatively lower accuracy is reasonable. Overall, our proposed method can still effectively differentiate the three types of long texts in general Chinese and English language data. In other words, long texts can be classified into these three types from the perspectives of coherence, cohesion, and complexity.

\begin{table*}[!t]
\centering\small
% \resizebox{\textwidth}{!}{%
\begin{tabular}{lccccc}
\toprule
& \textbf{Single-doc} & \textbf{Multi-doc} & \textbf{Sum} & \textbf{Few-shot} & \textbf{Synthetic} \\ 
\midrule
LLaMA2-7B-4K & 18.43 & 11.50 & 15.24 & 52.36 & \textbf{5.34} \\
HOL. + AGG. + CHA. & \textbf{23.71} & 12.54 & 17.32 & 59.23 & 3.45 \\
HOL.   & \underline{23.57} & \textbf{12.87} & 19.43 & 57.79 & 4.38 \\
HOL. + AGG. & 22.35 & 12.38 & \textbf{20.42} & \underline{59.68} & \underline{4.96}  \\
HOL. + Upsampling AGG. & 22.56 & \underline{12.74} & \underline{19.97} & \textbf{61.14} & 3.86 \\
\midrule
InternLM2-7B & \underline{43.50} & 37.10 & 23.70 & 59.95 & 40.33  \\
HOL. + AGG. + CHA. & 42.05 & 39.96 & 23.73 & 61.43 & 58.67 \\
HOL.   & 40.46 & \textbf{40.83} & 24.03 & \underline{62.07} & \underline{59.00} \\
HOL. + AGG. & 42.63 & \underline{40.35} & \underline{24.66} & 61.83 & 57.50 \\
HOL. + Upsampling AGG. & \textbf{44.20} & 40.15 & \textbf{25.28} & \textbf{62.70} & \textbf{63.05} \\ 
\bottomrule
\end{tabular}%
% }
\caption{Comparison of different training strategies data on the major task categories in LongBench. The terms HOL, AGG, and CHA respectively denote holistic texts, aggregated texts, and chaotic texts.}\label{tab_areas}
\end{table*}

\subsection{Setup}

We conduct experiments on LLaMA2-7B-4K~\citep{touvron2023llama2} and InternLM2-7B~\citep{internlm} corresponding to LLMs with and without long context capability respectively. Detailed training hyper-parameters can be found in Appendix~\ref{sec:hyper-parameters}.

For both LLaMA2-7B and InternLM2-7B, we use a 9:1 ratio of English to Chinese language data. 
For SlimPajama, we follow the data mixtures used for LLaMA pre-training~\citep{touvron2023llama}. Due to the limited amount of Chinese data, we sample data uniformly from Wanjuan. We excluded chaotic texts and upsample aggregated texts to balance the holistic and aggregated texts as our proposed recipe.

We compare our proposed data mixing recipe with the following three strategies:
1. Training on long texts from all categories.
2. Training LLM with only the holistic long texts. 
3. Excluding chaotic texts and employing holistic and aggregated texts for training.

\subsection{Main Results}

We first compare the training results of LLaMA2-7B and InternLM2-7B with our data mixing recipe mentioned above on LongWanjuan with other long-context LLMs, such as LongChat-v1.5-7B-32K~\cite{longchat}, Yi-6B-200K~\citep{Yi} and ChatGLM3-6B-32K~\citep{chatglm}, on LongBench~\citep{bai2023longbench}, a widely accepted benchmark dataset for long-context LLM. LongBench includes different languages (Chinese and English) and application areas (such as single-doc QA, multi-doc QA, summarization, few-shot learning tasks, synthetic tasks, and code completion) to provide a comprehensive evaluation of the language model's capabilities in handling long contexts. During the evaluation, we limit the maximum input length to 4K tokens for pre-trained LLaMA2-7B-4K and 32K tokens for other models. We apply the truncation from the middle used in LongBench. 

The results are shown in Table \ref{tab_main}, and detailed scores for each subtask can be found in the Appendix~\ref{sec:appendix_results}. 
Despite the strong long-text capabilities of InternLM2-7B, continuing training on LongWanjuan using our recipe leads to performance improvements across all domains. Moreover, we surpassed ChatGLM3-6B-32K overall, achieving a new state-of-the-art performance on LongBench.

\subsection{Analysis}
\label{sec:ablation}

Then we compare the training results of LLaMA2-7B and InternLM2-7B with the three strategies mentioned above. The results are shown in Table \ref{tab_long}, and detailed scores for each subtask can be found in Appendix \ref{sec:appendix_results}. Since our work mainly focuses on the quality of long text, we do not emphasize the improvement in code-related abilities. 
We observed that training solely on holistic texts yielded only marginal improvements compared to using data from all categories without any filtering. Incorporating aggregated texts leads to a slight decrease in performance for LLaMA-2 in the Chinese domain. When upsampling aggregated texts, both LLaMA-2 and InternLM-2 exhibits performance enhancements in both Chinese and English domains, achieving the optimal performance among these strategies.

We analyze the performance of these data mixing strategies across different tasks in Table~\ref{tab_areas}. For LLaMA2, the removal of chaotic texts results in improvements across multi-doc QA, summarization, few-shot learning tasks, and synthetic tasks. Additionally, incorporating aggregated texts alongside training solely on holistic texts enhances performance on these tasks. Although our proposed recipe excels primarily in few-shot learning tasks, it demonstrates overall superior performance.
Regarding InternLM2, our proposed recipe achieves optimal performance across all tasks except for multi-doc QA. We attribute the differing performances between the two models to the relatively lower proportion of Chinese in LLaMA2's pretraining corpus compared to our continued training with a 10\% Chinese ratio. Despite this distinction, our recipe yields the best overall performance on both these models.

We evaluate the performance of models fine-tuned on long texts across multiple short task benchmarks with a length of less than 2K tokens. Our findings indicate that the average performance fluctuation remains within 1.5 percentage points. Furthermore, incorporating aggregated texts proves to be effective in enhancing performance on short tasks. For detailed performance metrics and benchmark test results, please refer to the Appendix~\ref{sec:short_tasks}.

\section{Conclusion}

We try to systematically analyze the quality of long texts from three linguistic dimensions: coherence, cohesion, and complexity. Inspired by these dimensions, we develop a series of metrics based on statistics and pre-trained models to quantitatively assess the quality of long texts. Utilizing SlimPajama and Wanjuan, we construct the LongWanjuan dataset and categorize texts into three types: holistic, aggregated, and chaotic texts, according to our proposed metrics. We introduce a data mixture recipe based on the LongWanjuan dataset to address the issue of the imbalance between holistic long texts and aggregated long texts, achieving state-of-the-art performance on the LongBench benchmark. Our experimental analysis further validates the effectiveness of the proposed recipe.

% \section*{Acknowledgements}

\section*{Limitations}
We utilize SlimPajama and Wanjuan to construct LongWanjuan, with the Chinese data still remaining relatively limited. Based on the scalability and generalizability of our approach, additional Chinese datasets and datasets from other languages can be incorporated on top of deduplication. We alleviate the imbalance between the quantities of holistic and aggregated texts by upsampling aggregated texts. However, we did not attempt to provide an optimal ratio, leaving this for future work.

\section*{Ethics Statement}
LongWanjuan is constructed based on Wanjuan (under the CC BY 4.0 license) and SlimPajama (under the Apache 2.0 license), both of which permit open and free usage. We plan to open-source LongWanjuan under the CC BY 4.0 license. 

Throughout the dataset construction process, there are 3 annotators involved, all of whom are authors. The annotators are all native Chinese speaker and proficient in reading and understanding English. They consent to contribute their efforts to building LongWanjuan.

\bibliography{custom}

\newpage
\appendix

\section{Connectives and Pronouns}
\label{sec:conn}
The connectives and pronouns utilized in our metric calculations are outlined in Table~\ref{tab_conn} and Table~\ref{tab_pron}, respectively.

\begin{CJK*}{UTF8}{gbsn}
\begin{table*}[!t]
\centering{\small
\begin{tabular}{ll}
\toprule
Conn. in English & 'but ', 'whereas', 'however', 'though', 'yet', 'nevertheless', 'still', 'despite', \\
& 'nonetheless', 'notwithstanding', 'regardless of', 'in spite of', 'apart from',  \\
& 'in any case', 'in any event', 'supposedly', 'provided', 'otherwise', 'unless', 'once',  \\
& 'as long as', 'because', 'so ', 'since', 'thus', 'therefore', 'as a result',  \\
& 'accordingly', 'thereafter', 'thereby', 'hence', 'given', 'due to', 'owing to',  \\
& 'on account of', 'in light of', 'as a matter of fact', 'in other words', 'alternatively,', \\ 
& 'alternately,', 'optionally,', 'namely,', 'that is to say', 'in contrast', 'on the contrary', \\ 
& 'in turn', 'by contrast', 'conversely,', 'by comparison', 'for example', 'for instance', \\ 
& 'typically,', 'specifically,', 'especially,', 'particularly,', 'in particular', \\ 
& 'until', 'while', 'when', 'recently,', 'presently,', 'currently,', 'in the meantime', \\ 
& 'previously,', 'initially,', 'originally,', 'subsequently,', 'later', 'consequently,', \\ 
& 'finally,', 'ultimately,', 'eventually,', 'in the end', 'lately,', 'lastly,', \\ 
& 'firstly,', 'secondly,', 'thirdly,', 'next', 'on one hand', 'on the other hand', \\ 
& 'moreover', 'in addition', 'additionally,', 'besides', 'furthermore', \\ 
& 'in sum', 'in summary', 'overall', 'in short', 'in conclusion', 'in brief', 'in detail', \\ 
& 'personally,', 'luckily,', 'thankfully,', 'fortunately,', 'hopefully,', 'preferably,', \\ 
& 'surprisingly,', 'ironically,', 'amazingly,', 'oddly,', 'sadly,', 'historically,', \\ 
& 'traditionally,', 'theoretically,', 'practically,', 'realistically,', 'actually,', \\ 
& 'generally,', 'ideally,', 'technically,', 'honestly,', 'frankly,', 'basically,', \\ 
& 'admittedly,', 'undoubtedly,', 'importantly,', 'essentially,', 'naturally,', 'arguably,', \\ 
& 'remarkably,', 'in fact', 'in essence', 'in practice', 'in general', 'by doing this'. \\
\midrule
Conn. in Chinese & '至今为止，', '目前', '这样一来', '详细地', '与此同时，', '起初', '换言之', '此刻', \\ 
& '鉴于', '其中，', '例如，', '突然', '那么，', '不久，', '并且', '确实，', '尽管', \\ 
& '而不是', '总体上，', '第一，', '无论', '最近', '无论如何', '简而言之', '这里，', \\ 
& '有时候，', '除非', '结果，', '然后，', '除开', '当然，', '很快，', '但是，', \\ 
& '另一方面，', '换句话说，', '理论上', '历史上', '虽然', '不管', '所以，', \\ 
& '首先', '而且', '而', '由于', '第三，', '可是，', '但', '由此可见，', '而是', \\ 
& '最初，', '最终，', '后来，', '即使', '只有这样，', '但事实上，', '相反', \\ 
& '总的来说，', '只是', '取决于', '这时，', '用来', '以便', '基本上，', '不料', \\ 
& '就像', '接下来', '老实说', '相比之下，', '本质上', '否则，', '从某种意义上', \\ 
& '之前', '当时', '以前', '以至于', '特别是', '尤其是', '实际上，', '只要', \\ 
& '理想情况', '或者，', '不仅如此，', '幸运', '事实上，', '然而，', '一方面，', \\ 
& '比如，', '通常', '原因是', '从长远来看', '此后', '其次', '渐渐地，', '直到', \\ 
& '不论', '大多数情况下', '之后，', '显然', '也就是说，', '以及', '随后，', '没想到', \\ 
& '不过，', '除此之外', '无疑', '第二，', '反过来，', '若是', '以上就是', '也许', \\ 
& '假如', '可', '如果', '一如既往', '结果就是', '通过这样', '类似地，', '一般来说，', \\ 
& '除了', '据说', '另外，', '同样地', '反之，', '总之，', '进一步', '可以说', '于是，', \\ 
& '最后，', '既然', '尽管如此，', '这意味着', '同时，', '因此，', '某种程度上', \\ 
& '综上，', '随着', '此外，', '即便如此', '有时，', '同样，'. \\
\bottomrule
\end{tabular}}
\caption{The connectives we use to calculate  \(\text{Cohesion}_{\text{conn}}\). These words and phrases are collected from the list of connective words in \citet{dmr}.}\label{tab_conn}
\end{table*}
\end{CJK*}

\begin{CJK*}{UTF8}{gbsn}
\begin{table*}[!t]
\centering{\small
\begin{tabular}{ll}
\toprule
Pron. in English & 'one', 'ones', 'i', 'me', 'my', 'mine', 'myself', 'you', 'your', 'yours', 'yourself', \\
& 'he', 'him', 'his', 'himself', 'she', 'her', 'hers', 'herself', 'it', 'its', 'itself', \\
& 'we', 'us', 'our', 'ours', 'ourselves', 'they', 'them', 'their', 'theirs', 'themselves', \\
& 'this', 'that', 'these', 'those', 'who', 'whom', 'whose'. \\
\midrule
Pron. in Chinese & '我', '自己', '你', '他', '她', '它', '这', '那', '这个', '那个', '那里', '彼此', '您', \\
& '我们', '你们', '他们', '她们', '它们', '这些', '那些'. \\
\bottomrule
\end{tabular}}
\caption{The pronouns we use to calculate  \(\text{Cohesion}_{\text{pron}}\).}\label{tab_pron}
\end{table*}
\end{CJK*}

\section{Detailed Statistics}
\label{sec:detailed_stat}
We give an overview of the dataset statistics in the Chinese and English parts of LongWanjuan in Table~\ref{tab_en_statistics} and Table~\ref{tab_zh_statistics}, respectively.
\begin{table*}[!tb]
\centering\small
\begin{tabular}{lrrrrrrrr}
\toprule
\multirow{2}{*}{Domain} & \multicolumn{4}{c}{\#Docs}  & \multicolumn{4}{c}{\#Tokens} \\
 & Holistic & Aggregated & Chaotic & Total & Holistic & Aggregated & Chaotic & Total \\
\midrule
CommonCrawl	 & 4740880 & 638363 & 36664 & 5415907 & 76.5B & 9.9B & 719.8M & 87.2B \\
C4	 & 632819 & 88119 & 2732 & 723670 & 7.0B & 1.1B & 36.6M & 8.2B \\
ArXiv	 & 1045806 & 3274 & 287 & 1049367 & 25.4B & 153.9M & 68.3M & 25.6B \\
Book	 & 187396 & 7369 & 252 & 195017 & 24.2B & 893.9M & 80.7M & 25.1B \\
Github	 & 377312 & 56557 & 0 & 433869 & 7.4B & 1.3B & 0.0M & 8.7B \\
Wikipedia	 & 146469 & 29745 & 1883 & 178097 & 2.9B & 654.4M & 97.8M & 3.7B \\
StackExchange	 & 5295 & 1750 & 659 & 7704 & 60.6M & 21.9M & 11.3M & 93.8M \\
Total	 & 7234129 & 843211 & 48564 & 8125904 & 145.0B & 14.3B & 1.2B & 160.5B \\
\bottomrule
\end{tabular}
\caption{An overview of the dataset statistics in the English part of LongWanjuan. The number of tokens is calculated with the tokenizer in InternLM2-7B~\citep{internlm}.}\label{tab_en_statistics}
\end{table*}

\begin{table*}[!tb]
\centering\small
\begin{tabular}{lrrrrrrrr}
\toprule
\multirow{2}{*}{Domain} & \multicolumn{4}{c}{\#Docs}  & \multicolumn{4}{c}{\#Tokens} \\
 & Holistic & Aggregated & Chaotic & Total & Holistic & Aggregated & Chaotic & Total \\
\midrule
ChinaNews & 5211 & 1331 & 240 & 6782 & 51.3M & 15.5M & 4.3M & 71.1M \\
Law & 24575 & 5212 & 1310 & 31097 & 276.3M & 58.1M & 69.4M & 403.8M \\
Patent & 44922 & 2956 & 682 & 48560 & 438.0M & 31.6M & 9.9M & 479.5M \\
TextBook & 4746 & 693 & 0 & 5439 & 496.0M & 119.3M & 0.0M & 615.3M \\
WebText & 18698 & 7842 & 3855 & 30395 & 180.6M & 93.0M & 91.4M & 365.1M \\
Total & 98152 & 18034 & 6087 & 122273 & 1.4B & 317.4M & 175.1M & 1.9B \\
\bottomrule
\end{tabular}
\caption{An overview of the dataset statistics in the Chinese part of LongWanjuan. The number of tokens is calculated with the tokenizer in InternLM2-7B~\citep{internlm}.}\label{tab_zh_statistics}
\end{table*}

\section{Distribution of Texts across Metrics}\label{sec:text_dist}
\begin{figure}[h]
    \centering
    \includegraphics[width=1\linewidth]{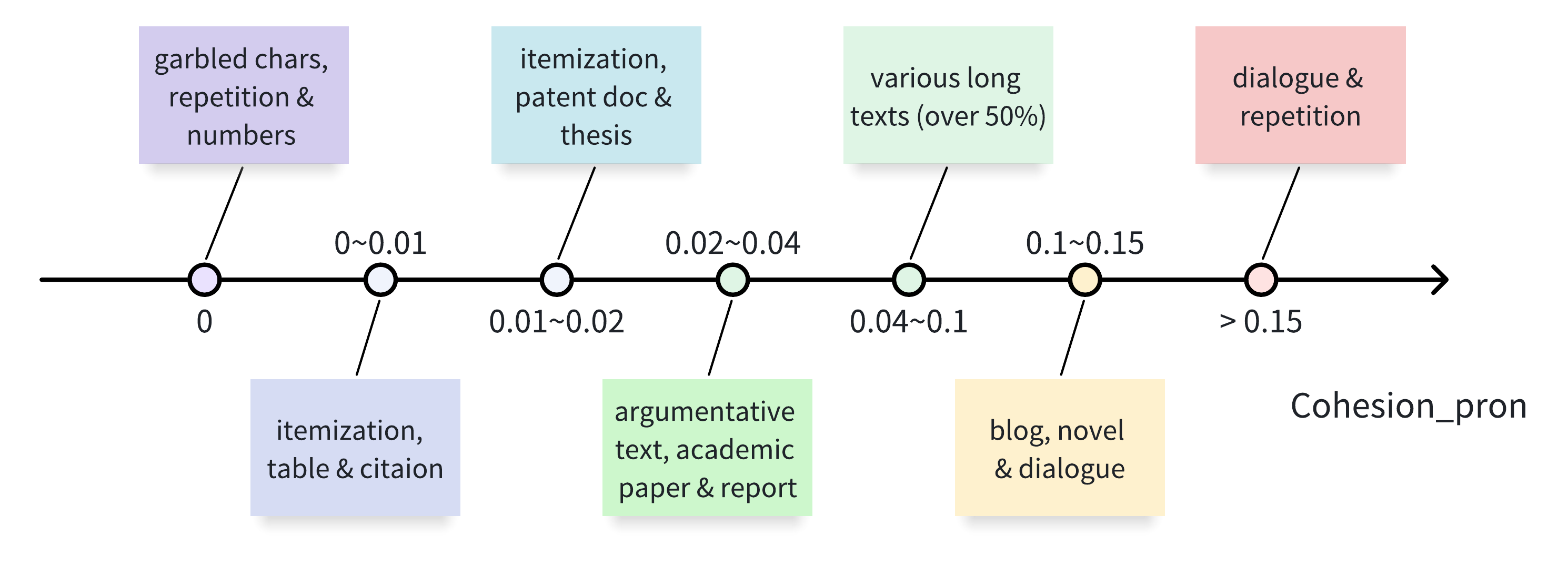}
    \caption{Distribution of texts with different characteristics on the \(\text{Cohesion}_\text{pron}\) metric in the C4 domain.}
    \label{fig:en_cohesion_pron}
\end{figure}

\begin{figure}[h]
    \centering
    \includegraphics[width=1\linewidth]{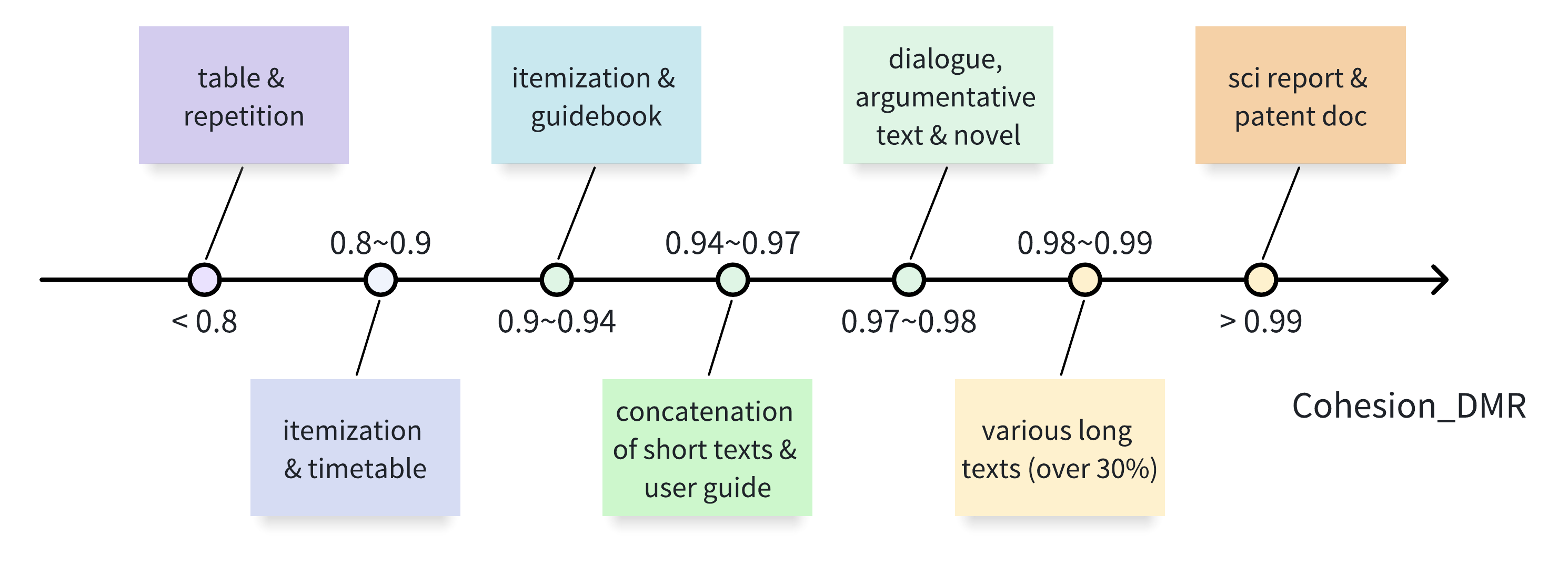}
    \caption{Distribution of texts with different characteristics on the \(\text{Cohesion}_\text{DMR}\) metric in the C4 domain.}
    \label{fig:en_cohesion_dmr}
\end{figure}

\begin{figure}[h]
    \centering
    \includegraphics[width=1\linewidth]{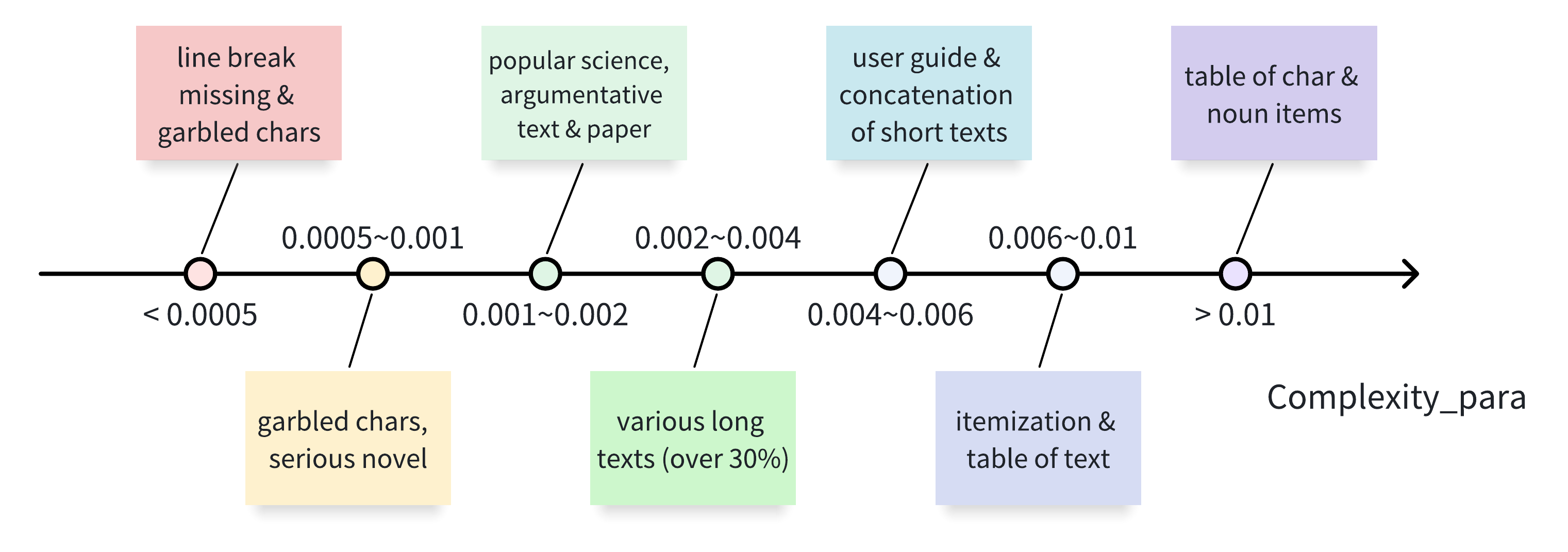}
    \caption{Distribution of texts with different characteristics on the \(\text{Complexity}_\text{para}\) metric in the C4 domain.}
    \label{fig:en_complexity_para}
\end{figure}

\begin{figure}[h]
    \centering
    \includegraphics[width=1\linewidth]{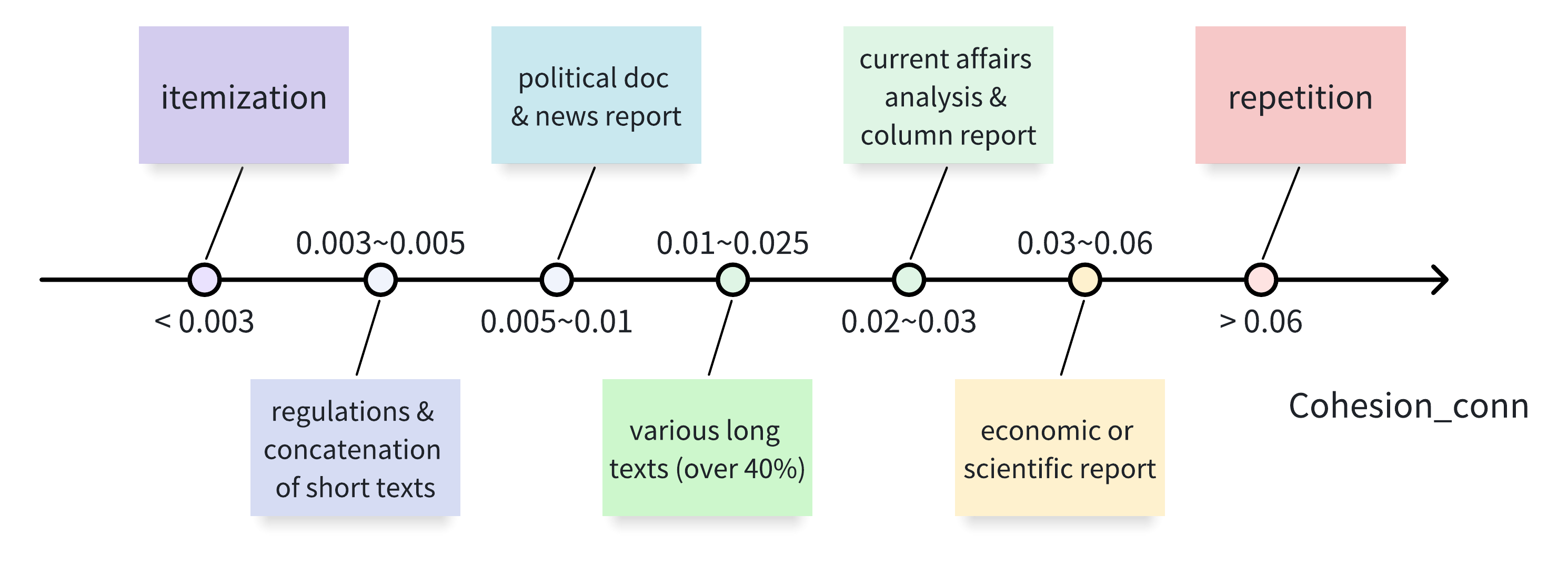}
    \caption{Distribution of texts with different characteristics on the \(\text{Cohesion}_\text{conn}\) metric in the ChinaNews domain.}
    \label{fig:zh_cohesion_conn}
\end{figure}

\begin{figure}[h]
    \centering
    \includegraphics[width=1\linewidth]{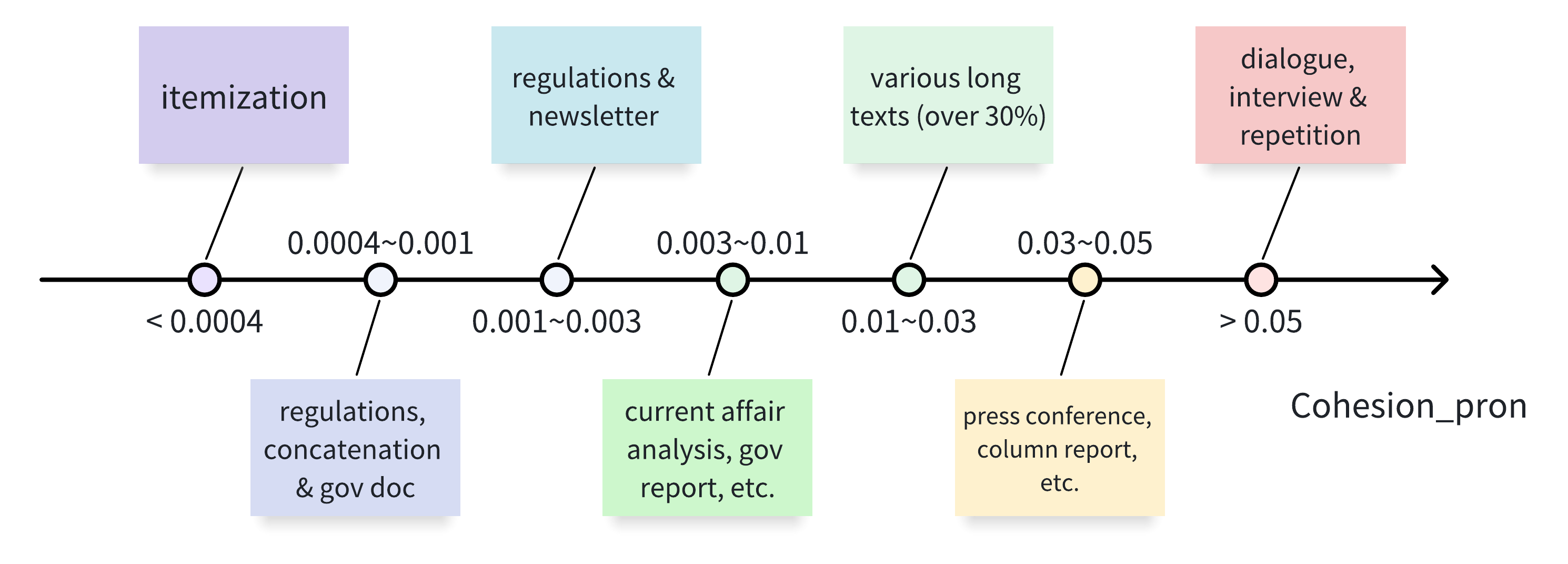}
    \caption{Distribution of texts with different characteristics on the \(\text{Cohesion}_\text{pron}\) metric in the ChinaNews domain.}
    \label{fig:zh_cohesion_pron}
\end{figure}

\begin{figure}[h]
    \centering
    \includegraphics[width=1\linewidth]{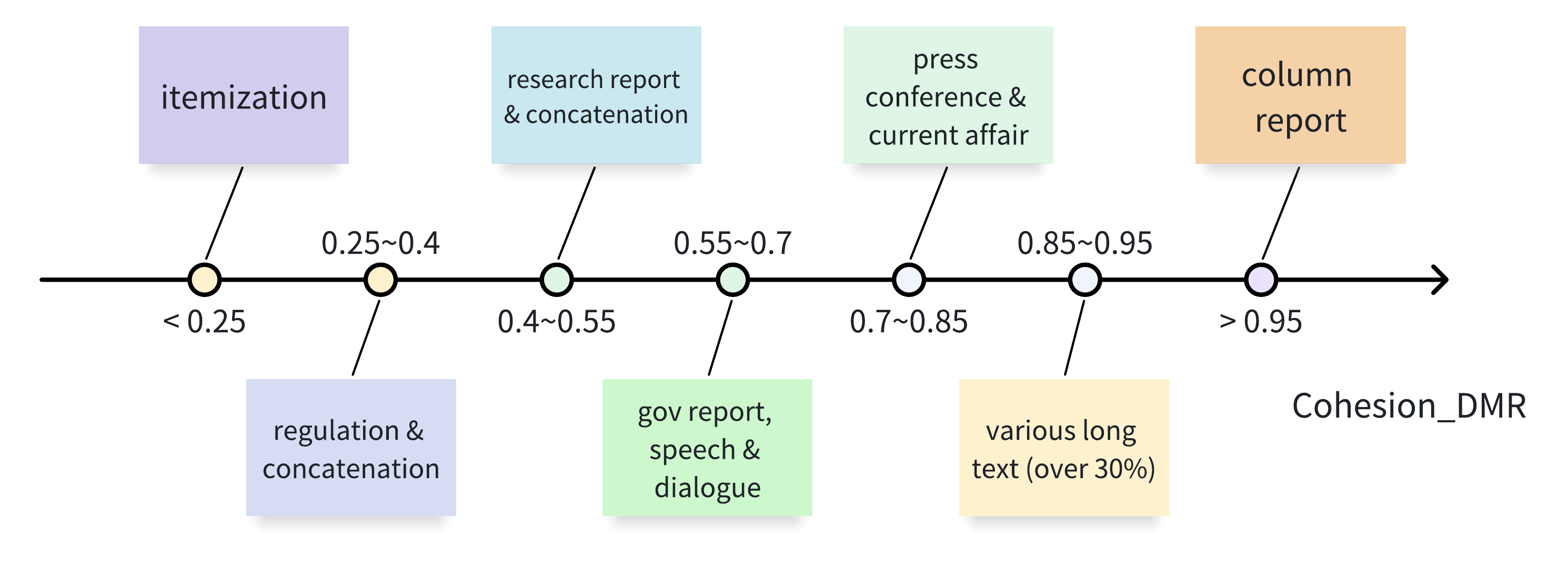}
    \caption{Distribution of texts with different characteristics on the \(\text{Cohesion}_\text{DMR}\) metric in the ChinaNews domain.}
    \label{fig:zh_cohesion_dmr}
\end{figure}

\begin{figure}[h]
    \centering
    \includegraphics[width=1\linewidth]{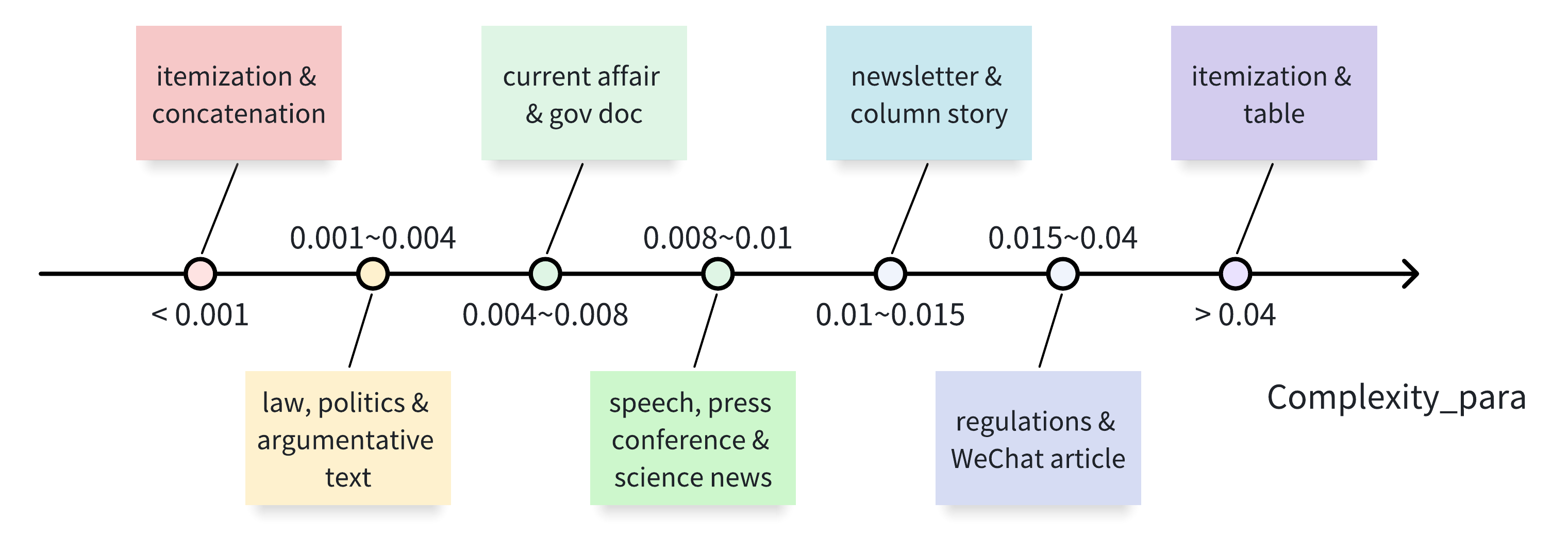}
    \caption{Distribution of texts with different characteristics on the \(\text{Complexity}_\text{para}\) metric in the ChinaNews domain.}
    \label{fig:zh_complexity_para}
\end{figure}

In this section, we report the distribution features with more characteristics, including \(\text{Cohesion}_\text{conn}\), \(\text{Cohesion}_\text{pron}\), \(\text{Cohesion}_\text{DMR}\), \(\text{Complexity}_\text{para}\), in Figure~\ref{fig:en_cohesion_pron} to Figure~\ref{fig:zh_complexity_para}. We take the C4 domain and the ChinaNews domain as an example of English and Chinese texts respectively.

% \begin{table*}[!tb]
% \centering\small
% \begin{tabular}{@{}lllll@{}}
% \toprule
% \multirow{2}{*}{Domain} & Size             & Holistic         & Aggregated       & Chaotic         \\ \cmidrule(l){2-5} 
%                         & \multicolumn{4}{c}{Tokens / Docs}                                        \\ \midrule
% ChinaNews               & 71.07M / 6.78K   & 51.27M / 5.21K   & 15.48M / 1.33K   & 4.33M / 0.24K   \\
% Law                     & 403.78M / 31.10K & 276.26M / 24.57K & 58.09M / 5.21K   & 69.44M / 1.31K  \\
% Patent                  & 479.46M / 48.56K & 437.96M / 44.92K & 31.57M / 2.96K   & 9.92M / 0.68K   \\
% TextBook                & 615.27M / 5.44K  & 496.01M / 4.75K  & 119.26M / 0.69K  & 0.00K / 0.00K   \\
% WebText                 & 365.06M / 30.39K & 180.60M / 18.70K & 93.04M / 7.84K   & 91.43M / 3.85K  \\
% Total                   & 1.93B / 122.27K  & 1.44B / 98.15K   & 317.45M / 18.03K & 175.11M / 6.09K \\ \bottomrule
% \end{tabular}
% \caption{zh_statistics}\label{tab_zh_statistics}
% \end{table*}

\section{Detailed Results}
\label{sec:appendix_results}
Detailed results of all the models we tested are shown in Table~\ref{tab_longbench_qa}, Table~\ref{tab_longbench_sum_icl} and Table~\ref{tab_longbench_syn_code}.
\begin{table*}[!t]
\centering{\small
\begin{tabular}{lcccccccc}
\toprule
& \textbf{Narrative} & \textbf{Qasper} & \textbf{MF\_en} & \textbf{MF\_zh} & \textbf{Hotpot} & \textbf{2Wikim} & \textbf{Musique} & \textbf{Dureader} \\
&\textbf{QA}&&&&\textbf{QA}&\textbf{QA}&&\\
\midrule
LLaMA2-7B-4K & 16.86 & 15.35 & 23.78 & 19.08 & 7.85 & 10.54 & 4.27 & 23.34 \\
HOL. + AGG. + CHA. & 22.61 & 20.39 & 30.60 & 22.96 & 9.34 & 10.78 & 6.01 & 24.01 \\
HOL. & 15.36 & 19.12 & 35.04 & 27.64 & 9.74 & 10.83 & 6.00 & 24.89 \\
HOL. + AGG. & 19.15 & 19.68 & 29.60 & 22.78 & 10.36 & 10.49 & 5.47 & 23.19 \\
HOL. + Upsampling AGG. & 16.93 & 20.16 & 26.43 & 27.68 & 9.63 & 10.82 & 6.75 & 23.77 \\
\midrule
InternLM2-7B & 24.02 & 41.97 & 47.95 & 61.16 & 52.98 & 37.89 & 28.02 & 29.52 \\
HOL. + AGG. + CHA. & 26.86 & 39.95 & 41.28 & 59.90 & 54.76 & 43.03 & 31.04 & 31.00 \\
HOL. & 22.52 & 40.46 & 39.99 & 58.76 & 54.77 & 45.07 & 32.28 & 31.18 \\
HOL. + AGG. & 27.25 & 40.29 & 42.92 & 60.14 & 53.75 & 44.53 & 30.87 & 32.25 \\
HOL. + Upsampling AGG. & 29.93 & 39.62 & 50.17 & 58.57 & 53.68 & 42.31 & 32.14 & 32.46 \\
\midrule
LongChat-v1.5-7B-32K & 16.90 & 27.70 & 41.40 & 29.10 & 31.50 & 20.60 & 9.70 & 19.50 \\
Yi-6B-200K & 12.36 & 26.41 & 36.78 & 22.36 & 46.57 & 40.38 & 25.78 & 14.73 \\
ChatGLM3-6B-32K & 9.21 & 43.07 & 50.86 & 60.33 & 55.33 & 43.73 & 38.94 & 41.89 \\
\bottomrule
\end{tabular}}
\caption{Results on single-doc and multi-doc QA subtasks in Longbench including NarrativeQA, Qasper, MultiField\_en (MF\_en), MultiField\_zh (MF\_zh), HotpotQA, 2WikimQA, Musique, and Dureader.}
\label{tab_longbench_qa}
\end{table*}

\begin{table*}[!t]
\centering{\small
\begin{tabular}{lcccccccc}
\toprule
& \textbf{Gov} & \textbf{QMSum} & \textbf{MultiNews} & \textbf{VCSum} & \textbf{TREC} & \textbf{Trivia} & \textbf{SAM} & \textbf{LSHT} \\
&\textbf{Report}&&&&&\textbf{QA}&\textbf{Sum}&\\
\midrule
LLaMA2-7B-4K & 27.09 & 20.63 & 3.21 & 10.02 & 68.00 & 89.09 & 32.09 & 20.25 \\ 
HOL. + AGG. + CHA. & 29.54 & 21.75 & 6.61 & 11.37 & 70.00 & 86.75 & 39.15 & 41.00 \\ 
HOL. & 28.66 & 21.35 & 16.34 & 11.36 & 69.00 & 88.44 & 32.71 & 41.00 \\ 
HOL. + AGG. & 30.72 & 21.58 & 18.26 & 11.11 & 71.00 & 88.36 & 39.36 & 40.00 \\
HOL. + Upsampling AGG. & 28.87 & 22.14 & 16.46 & 12.42 & 71.50 & 88.78 & 39.78 & 44.50 \\ 
\midrule
InternLM2-7B & 30.02 & 23.09 & 26.46 & 15.23 & 75.50 & 92.36 & 30.94 & 41.00 \\ 
HOL. + AGG. + CHA. & 33.69 & 25.03 & 27.14 & 9.05 & 76.00 & 89.41 & 37.99 & 42.33 \\ 
HOL. & 33.68 & 25.29 & 27.04 & 10.12 & 77.00 & 89.17 & 38.85 & 43.25 \\ 
HOL. + AGG. & 33.49 & 25.64 & 27.54 & 11.95 & 77.00 & 89.07 & 37.43 & 43.83 \\
HOL. + Upsampling AGG. & 32.96 & 25.49 & 27.84 & 14.81 & 77.00 & 91.29 & 41.00 & 41.50 \\ 
\midrule
LongChat-v1.5-7B-32K & 30.80 & 22.70 & 26.40 & 9.90 & 63.50 & 82.30 & 34.20 & 23.20 \\
Yi-6B-200K & 29.34 & 20.65 & 27.14 & 8.14 & 73.50 & 86.94 & 9.85 & 37.50 \\
ChatGLM3-6B-32K & 35.99 & 24.68 & 27.44 & 15.83 & 79.00 & 87.39 & 17.72 & 42.00 \\ 
\bottomrule
\end{tabular}}
\caption{Results on summarization and few-shot learning subtasks in Longbench including GovReport, QMSum, MultiNews, VCSum, TREC, TriviaQA, SAMSum, and LSHT.}
\label{tab_longbench_sum_icl}
\end{table*}

\begin{table*}[!t]
\centering{\small
\begin{tabular}{lccccc}
\toprule
& \textbf{PC} & \textbf{PR\_en} & \textbf{PR\_zh} & \textbf{LCC} & \textbf{Repobench-p} \\
\midrule
LLaMA2-7B-4K & 1.50 & 5.52 & 9.00 & 68.22 & 62.25 \\
HOL. + AGG. + CHA. & 2.05 & 4.55 & 3.75 & 65.17 & 60.91 \\ 
HOL. & 2.00 & 5.38 & 5.75 & 65.97 & 61.33  \\ 
HOL. + AGG. & 1.50 & 7.62 & 5.75 & 65.10 & 60.52 \\
HOL. + Upsampling AGG. & 2.50 & 3.82 & 5.25 & 65.93 & 59.86 \\ 
\midrule
InternLM2-7B & 7.00 & 56.50 & 57.50 & 63.90 & 61.81  \\ 
HOL. + AGG. + CHA. & 2.00 & 96.50 & 77.50 & 69.96 & 64.58 \\ 
HOL. & 0.00 & 98.50 & 78.50 & 69.42 & 65.39 \\ 
HOL. + AGG. & 0.50 & 96.00 & 76.00 & 69.13 & 65.06 \\
HOL. + Upsampling AGG. & 3.14 & 97.50 & 88.50 & 66.80 & 63.71 \\ 
\midrule
LongChat-v1.5-7B-32K & 1.00 & 30.50 & 7.60 & 53.00 & 55.30 \\
Yi-6B-200K & 2.50 & 6.00 & 7.97 & 66.10 & 63.00 \\
ChatGLM3-6B-32K & 2.00 & 98.50 & 94.50 & 60.07 & 54.12 \\ 
\bottomrule
\end{tabular}}
\caption{Results on synthetic and code subtasks in Longbench including PassageCount (PC), PassageRetrieval\_en (PR\_en), PassageRetrieval\_zh (PR\_zh), LCC and Repobench-p.}
\label{tab_longbench_syn_code}
\end{table*}

\section{Hyper-parameters}
\label{sec:hyper-parameters}
We use 64 A100 GPUs and adopt ZeRO3 strategies~\citep{rajbhandari2020zero} to tune a 7B model. We use AdamW \citep{loshchilov2017decoupled} with $\beta_1=0.9$ and $\beta_2=0.95$. We set the learning rate to $3\times10^{-5}$ with a cosine learning rate schedule with a 20-step warmup. We set the max gradient norm to 1 and the weight decay to zero. 

We fine-tune both LLaMA2-7B-4K and InternLM2-7B with 5B tokens using the next token prediction objective. We set the global batch size to 2M tokens, with a max length of 32K tokens. 
Specifically, for the fine-tuning of LLaMA2-7B to achieve context over 32K tokens, we adjust the base of the rotation angle in RoPE~\citep{su2024roformer} to 500000 based on LLaMA2Long~\citep{llama2long} and ScalingRoPE~\citep{ScalingRope}.

\section{Performance on Short Tasks}
\label{sec:short_tasks}
To verify that the LLM trained on long text in our proposed strategies can still achieve good performance on short-text tasks, we also evaluate our fine-tuned LLaMA2-7B and InternLM2-7B with a maximum input context of 2K tokens on short tasks, including ARC-easy/challenge~\citep{arc}, Hellaswag~\citep{zellers2019hellaswag}, Winogrande~\citep{sakaguchi2021winogrande}, TruthfulQA~\citep{DBLP:conf/acl/LinHE22}, SuperGLUE~\citep{wang2019superglue}, GSM8K~\citep{gsm8k} and MMLU~\citep{mmlu}. The results are shown in Table \ref{tab_short}.
\begin{table*}[!t]
\centering\small
\begin{tabular}{lccccccccc}
\toprule
& \textbf{GSM8K} & \textbf{ARC-e} & \textbf{ARC-c} & \textbf{HS} & \textbf{WG} & \textbf{TQA} & \textbf{SG} & \textbf{MMLU} & \textbf{Average} \\
\midrule
LLaMA2-7B-4K & 16.30 & 52.73 & 36.95 & 69.24 & 61.25 & 35.09 & 50.43 & 46.78 & 46.10 \\
HOL. + AGG. + CHA. & 16.45 & 53.09 & 34.24 & 65.11 & 61.01 & 36.11 & 51.25 & 44.13 & 45.17 \\
HOL. & 15.54 & 53.09 & 33.90 & 65.46 & 61.40 & 34.80 & 51.40 & 42.71 & 44.79 \\
HOL. + AGG. & 16.76 & 54.67 & 35.93 & 65.90 & 61.01  & 36.40 & 50.60 & 44.74 & 45.75 \\
HOL. + Upsampling AGG. & 17.13 & 53.97 & 33.22 & 65.86 & 60.30 & 36.26 & 49.50 & 44.49 & 45.09 \\
\midrule
InternLM2-7B & 69.83 & 51.50 & 42.37 & 54.87 & 77.35 & 39.62 & 78.83 & 65.60 & 60.00 \\
HOL. + AGG. + CHA. & 69.67 & 58.38 & 41.69 & 64.46	& 78.93 & 37.43 & 78.43 & 64.45 & 61.68 \\
HOL. & 70.20 & 50.26 & 42.37 & 56.87	& 77.90 & 38.30 & 79.01 & 64.75 & 59.96 \\
HOL. + AGG. & 70.43 & 55.56 & 40.34 & 61.64 & 77.43 & 37.57 & 78.85 & 64.11 & 60.74 \\
HOL. + Upsampling AGG. & 68.99 & 57.14 & 41.69 & 65.46 & 78.61 & 38.30 & 79.20 & 64.11 & 61.69 \\ 
\bottomrule
\end{tabular}
\caption{Results on 0-shot ARC-easy/challenge, Hellaswag (HS), Winogrande (WG), TruthfulQA (TQA), SuperGLUE (SG), 4-shot GSM8K and 5-shot MMLU.}
\label{tab_short}
\end{table*}

\end{document}